\documentclass[10pt,twocolumn,letterpaper]{article}

\usepackage{wacv}              

\usepackage{graphicx}
\usepackage{amsmath}
\usepackage{amssymb}
\usepackage{booktabs}
\usepackage{xcolor}
\usepackage{adjustbox}
\usepackage{multicol}
\usepackage{url}
\usepackage{graphicx}
\usepackage{amsmath}
\usepackage{amssymb}
\usepackage{multicol}
\usepackage{makecell}
\usepackage{subcaption}
\usepackage{blindtext}
\usepackage{cite}
\usepackage{tabularx}

\usepackage{latexsym}
\usepackage{graphicx}
\usepackage{textcomp}

\usepackage{framed,multirow}

\usepackage{amsmath,amssymb,amsfonts}
\usepackage{algorithmic}
\usepackage{times}
\usepackage{epsfig}
\usepackage[accsupp]{axessibility}

\usepackage[pagebackref,breaklinks,colorlinks]{hyperref}

\usepackage[capitalize]{cleveref}
\crefname{section}{Sec.}{Secs.}
\Crefname{section}{Section}{Sections}
\Crefname{table}{Table}{Tables}
\crefname{table}{Tab.}{Tabs.}

\def\wacvPaperID{513} 
\def\confName{WACV}
\def\confYear{2024}

\begin{document}

\title{DTrOCR: Decoder-only Transformer for Optical Character Recognition}

\author{Masato Fujitake\\
FA Research, 
Fast Accounting Co., Ltd.
Japan \\
{\tt\small fujitake@fastaccounting.co.jp}
}
\maketitle

\begin{abstract}
    Typical text recognition methods rely on an encoder-decoder structure, in which the encoder extracts features from an image, and the decoder produces recognized text from these features. 
In this study, we propose a simpler and more effective method for text recognition, known as the Decoder-only Transformer for Optical Character Recognition (DTrOCR).
This method uses a decoder-only Transformer to take advantage of a generative language model that is pre-trained on a large corpus.
We examined whether a generative language model that has been successful in natural language processing can also be effective for text recognition in computer vision.  
Our experiments demonstrated that DTrOCR outperforms current state-of-the-art methods by a large margin in the recognition of printed, handwritten, and scene text in both English and Chinese.
\end{abstract}


\section{Introduction} \label{intro}
The aim of text recognition, also known as optical character recognition (OCR), is to convert the text in images into digital text sequences.
Many studies have been conducted on this technology owing to its wide range of real-world applications, including reading license plates and handwritten text, analyzing documents such as receipts and invoices~\cite{huang2019sroie, yiheng2020layoutlm}, and analyzing road signs in automated driving and natural scenes~\cite{fujitake2021tcbam, fujitake2023a3s}.
However, the various fonts, lighting variations, complex backgrounds, low-quality images, occlusion, and text deformation make text recognition challenging.
Numerous methods have been proposed to overcome these challenges.

\begin{figure}[htp]
	\centering
	\includegraphics[width=1.00\columnwidth, keepaspectratio]{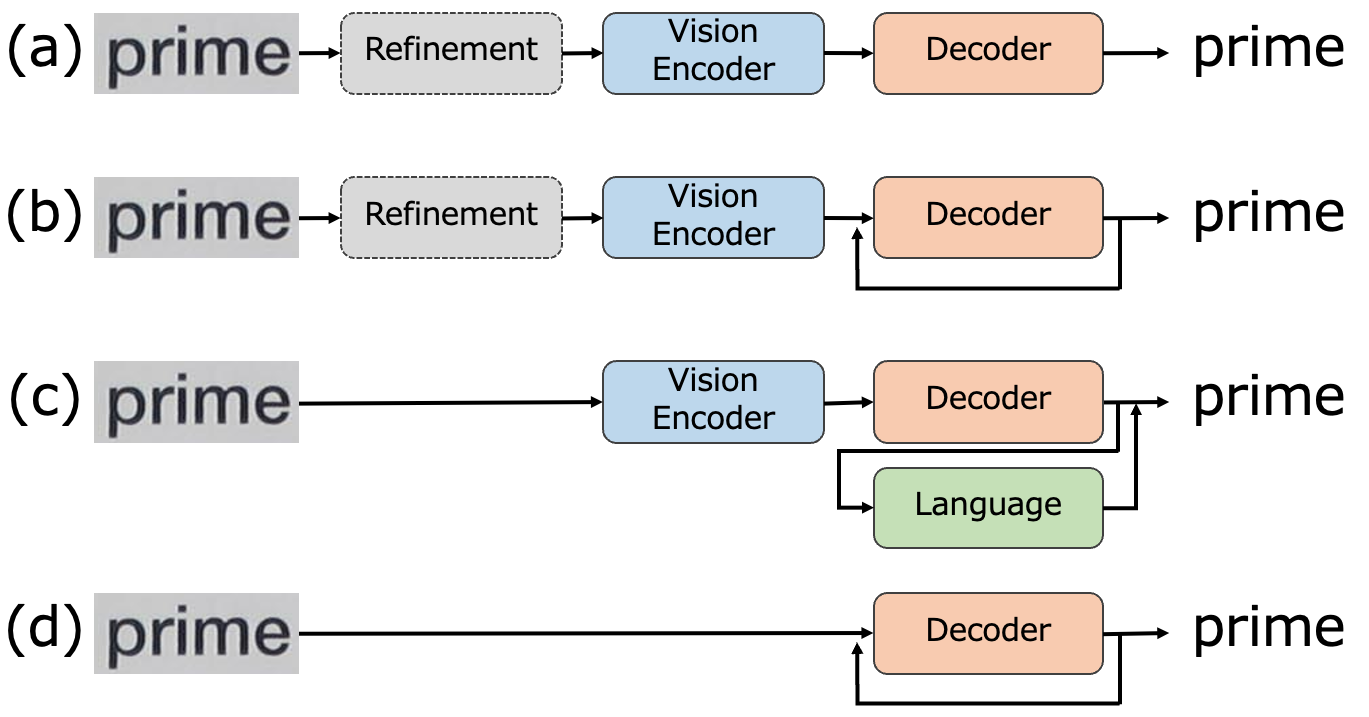}
	\caption{
Model patterns for text recognition.
The existing methods in (a) to (c) consist of a vision encoder to extract the image features and a decoder to predict text sequences from the features.
Some methods use refinement modules to deal with low-quality images and an LM to correct the output text.
Our approach differs significantly.
As shown in (d), it consists of a simple model pattern in which the image is fed directly into the decoder to generate text.
 }
	\label{fig:several_arch}
\end{figure}

Existing approaches have mainly employed an encoder-decoder architecture for robust text recognition~\cite{fang2021ABINet, shi2016crnn, bautista2022parseq}.
In such methods, the encoder extracts the intermediate features from the image, and the decoder predicts the corresponding text sequence.
Figures~\ref{fig:several_arch} (a)--(c) present the encoder-decoder model patterns of previous studies.
The methods in Figures~\ref{fig:several_arch} (a) and (b) employ the convolutional neural network (CNN)~\cite{he2016resnet} and Vision Transformer (ViT)~\cite{dosovitskiy2020vit} families as image encoding methods, with the recurrent neural network (RNN)~\cite{hochreiter1997lstm} and Transformer~\cite{vaswani2017transformer} families as decoders, which can be used for batch inference (Figure~\ref{fig:several_arch} (a)) or recursively inferring characters one by one (Figure~\ref{fig:several_arch} (b)).
Certain modules apply image curve correction~\cite{baek2021TRBA} and high-resolution enhancement~\cite{mou2020plugnet} to the input images to boost the accuracy.
Owing to the limited information from images, several methods in recent years have focused on linguistic information and have employed language models (LMs) in the decoder either externally~\cite{fang2021ABINet} or internally~\cite{bautista2022parseq, li2021trocr}, as illustrated in Figure~\ref{fig:several_arch} (c).
Although these approaches can achieve high accuracy by leveraging linguistic information, the additional computational cost is an issue.

LMs based on generative pre-training have been used successfully in various natural language processing (NLP) tasks~\cite{radford2018gpt, radford2019language} in recent years.
These models are built with a decoder-only Transformer. 
The model passes the input text directly to the decoder, which outputs the subsequent word token.
The model is pre-trained with a large corpus to generate the following text sequence in the given text.
The model needs to understand the meaning and context of the words and acquire language knowledge to generate the text sequence accurately.
As the obtained linguistic information is powerful, it can be fine-tuned for various tasks in NLP.
However, the applicability of such models to text recognition has yet to be demonstrated.

Motivated by the above observations, this study presents a novel text recognition framework known as the Decoder-only Transformer for Optical Character Recognition (DTrOCR), which does not require a vision encoder for feature extraction.
The proposed method transforms a pre-trained generative LM with a high language representation capability into a text recognition model.
Generative models that are used in NLP use a text sequence as input and generate the following text autoregressively.
In this work, the model is trained to use an image as the input. 
The input image is converted into a patch sequence, and the recognition results are output autoregressively.
The structure of the proposed method is depicted in Figure~\ref{fig:several_arch} (d).
DTrOCR does not require vision encoders such as a CNN or ViT but has a simple model structure with only a decoder that leverages the internal generative LM.
In addition, the proposed method employs fine-tuning from a pre-trained model, which reduces the computational resources.
Despite its simple structure, DTrOCR is revealed to outperform existing methods in various text recognition benchmarks, such as handwritten, printed, and natural scene image text, in English and Chinese.
The main contributions of this work are summarized as follows:
\begin{itemize}
    \item We propose a novel decoder-only text recognition method known as DTrOCR, which differs from the mainstream encoder-decoder approach.
    \item Despite its simple structure, DTrOCR achieves state-of-the-art results on various benchmarks by leveraging the internal LM, without relying on complex pre- or post-processing.

\end{itemize}

\section{Related Works}
\noindent
\textbf{Scene Text Recognition.}
Scene text recognition refers to text recognition in natural scene images.
Existing methods can be divided into three categories: word-based, character-based, and sequence-based approaches.
Word-based approaches perform text recognition as image classification, in which each word is a direct classification class~\cite{jaderberg2016MJSynth}.
Character-based approaches perform text recognition using detection, recognition, and grouping on a character-by-character basis~\cite{wan2020textscanner}.
Sequence-based approaches deal with the task as sequence labeling and are mainly realized using encoder-decoder structures~\cite{shi2016crnn, fang2021ABINet, li2021trocr}.
The encoder, which can be constructed using the CNN and ViT families, aims to extract a visual representation of a text image.
The goal of the decoder is to map the representation to text with connectionist temporal classification (CTC)-based methods~\cite{graves2006ctcloss, shi2016crnn} or attention mechanisms~\cite{bautista2022parseq, li2021trocr, yu2020srn, lu2021master}.
The encoder has been improved using a neural architecture search~\cite{zhang2020autostr} and a graph convolutional network~\cite{yan2021pren2d}, whereas the decoder has been enhanced using multistep reasoning~\cite{bhunia2021joint}, two-dimensional features~\cite{li2019show}, semantic learning~\cite{qiao2020seed}, and feedback~\cite{bhunia2021towards}.
Furthermore, the accuracy can be improved by converting low-resolution inputs into high-resolution inputs~\cite{mou2020plugnet}, normalizing curved and irregular text images~\cite{shi2018aster, luo2019moran, baek2021TRBA}, or using diffusion models~\cite{fujitake2023diffusionstr}.
Some methods have predicted the text directly from encoders alone for computational efficiency~\cite{du2022svtr, atienza2021vitstr}.
However, these approaches do not use linguistic information and face difficulties when the characters are hidden or unclear.

Therefore, methods that leverage language knowledge have been proposed to make the models more robust in recent years.
ABINet uses bi-directional context via external LMs~\cite{fang2021ABINet}.
In VisionLAN, an internal LM is constructed by selectively masking the image features of individual characters during training~\cite{wang2021visionlan}.
The learning of internal LMs using permutation language modeling was proposed in PARSeq~\cite{bautista2022parseq}.
In TrOCR, LMs that are pre-trained on an NLP corpus using masked language modeling (MLM) are used as the decoder~\cite{li2021trocr}.
MaskOCR includes sophisticated MLM pre-training methods to enhance the Transformer-based encoder-decoder structure~\cite{lyu2022maskocr}.
The outputs of these encoders are either passed directly to the decoder or intricately linked by a cross-attention mechanism.

Our approach exhibits similarities to TrOCR~\cite{li2021trocr} as both use linguistic information and pre-trained LMs for the decoding process.
However, our method differs in two significant aspects. 
First, we use generative pre-training~\cite{radford2018gpt} as the pre-training method in the decoder, which predicts the next word token for generating text, instead of solving masked fill-in-the-blank problems using MLM. 
Second, we eliminate the encoder to obtain elaborate features from images.
The images are patched and directly fed into the decoder. 
As a result, no complicated connections such as cross-attention are required to link the encoder and decoder because the image and text information are handled at the same sequence level. 
This enables our text recognition model to be simple yet effective.

\noindent
\textbf{Handwritten Text Recognition.}
Handwritten text recognition (HWR) has long been studied, and the recent methods have been reviewed~\cite{memon2020hwr_review}. 
In addition, the effects of different attention mechanisms of encoder-decoder structures in the HWR domain have been compared~\cite{michael2019evaluating}.
The combination of RNNs and CTC decoders has been established as the primary approach in this field~\cite{bluche2017gated, michael2019evaluating, wang2020decoupled}, with improvements such as multi-dimensional long short-term memory~\cite{puigcerver2017multidimensional} and attention mechanisms~\cite{kang2022pay, diaz2021rethinking} having been applied.
In recent years, extensions using LMs have also been implemented~\cite{li2021trocr}.
Thus, we tested the effectiveness of our method in HWR to confirm its scalability.

\noindent
\textbf{Chinese Text Recognition.}
Text recognition tasks on alphabets and symbols in English have been studied, and significant accuracy improvements have been achieved~\cite{fang2021ABINet, bautista2022parseq}.
The adaptation of recognition models to Chinese text recognition (CTR) has been investigated in recent years~\cite{yu2021btcr, lyu2022maskocr}.
However, studies on CTR remain lacking.
CTR is a challenging task as Chinese has substantially more characters than English and contains many similar-appearing characters.
Thus, we validated our method to determine whether it can be applied to Chinese in addition to English text.

\noindent
\textbf{Generative Pre-Trained Transformer.}
Generative pre-trained Transformer (GPT) has emerged in NLP, which has attracted attention owing to its ability to produce results in various tasks~\cite{radford2018gpt, radford2019language}. 
The model can acquire linguistic ability by predicting the continuation of a given text.
GPT comprises a decoder-only autoregressive transformer that does not require an encoder to acquire the input text features.
Whereas previous studies~\cite{li2021trocr} examined the adaptation of LMs that are learned using MLM to text recognition models, this study explores the extension of GPT to text recognition.

\section{Method} \label{method}
\begin{figure*}[htp]
	\centering
	\includegraphics[width=1.70\columnwidth, keepaspectratio]{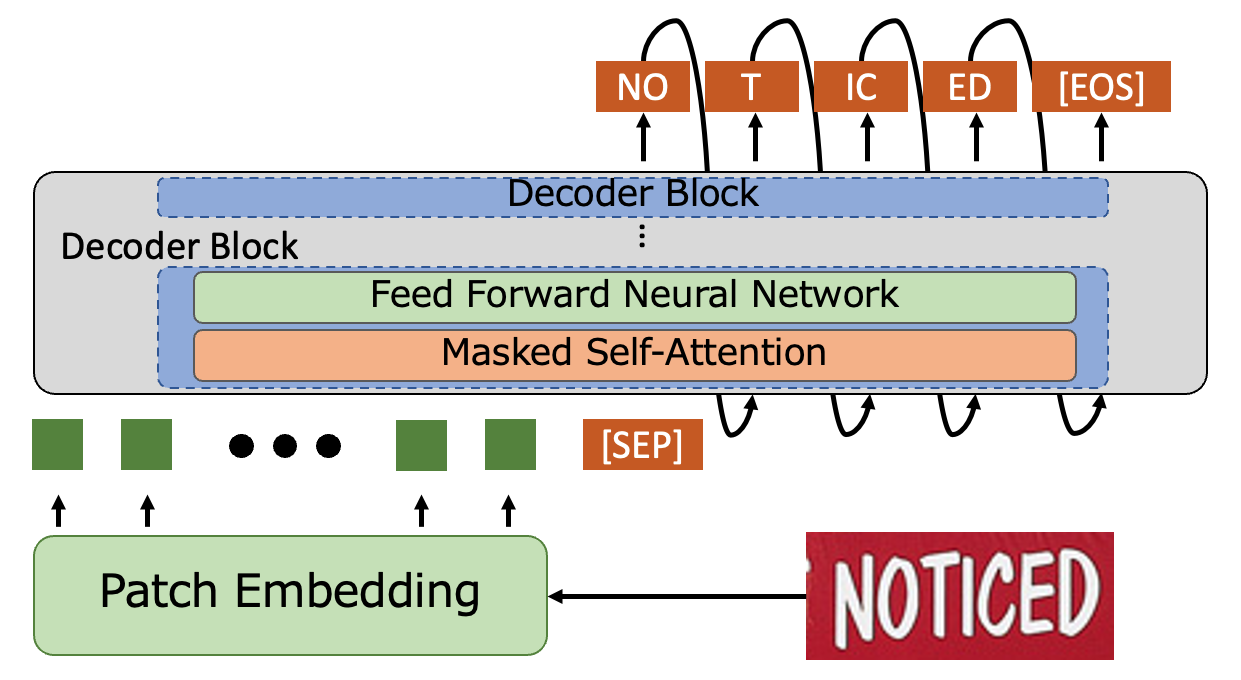}
	\caption{
Architecture of proposed DTrOCR, which
consists of patch embedding and decoder modules.
The input images are transformed into one-dimensional sequences using the patch embedding and then sent to the decoder along with the positional encoding. 
The decoder uses the special token \texttt{[SEP]} to indicate sequence separation.
Thereafter, it predicts the subsequent word token based on the sequence condition.
It continues to generate text autoregressively until it reaches the end of the text token \texttt{[EOS]}.
	}
	\label{fig:overall_architecture}
\end{figure*}

The pipeline of our method is illustrated in Figure~\ref{fig:overall_architecture}.
We use a generative LM~\cite{radford2019language} that incorporates Transformers~\cite{vaswani2017transformer} with a self-attention mechanism. 
Our model comprises two main components: the patch embedding and Transformer decoder. 
The input text image undergoes patch embedding to divide it into patch-sized image sequences. 
Subsequently, these sequences are passed through the decoder with the positional embedding. 
The decoder predicts the first-word token after receiving a special token \texttt{[SEP]} that separates the image and text sequence. 
Thereafter, it predicts the subsequent tokens in an autoregressive manner until the special token \texttt{[EOS]}, which indicates the end of the text, and produces the final result. 
The modules and training methods are described in detail in the following sections.

\subsection{Patch Embedding Module}
Since the input to the transformer is a sequence of tokens, the patch embedding as the input to the Transformer is a sequence of tokens, the patch embedding module transforms the tokens so that the image can be input into the decoder.
We employ the patch embedding procedure proposed in~\cite{dosovitskiy2020vit}.
The input image is resized to a fixed-size image $I \in \mathbb{R}^{W \times H \times C}$, where $W$, $H$, and $C$ are the width, height, and channel of the image, respectively.
The input image is divided by fixed patch sizes $p_{w} \times p_{h}$, where $p_{w}$ and $p_{h}$ are the width and height of the patch, respectively.
The patch images are first transformed into vectors and adjusted to fit the input dimensions of the Transformer.
Position encoding is added to preserve the information on the position of each patch. 
Subsequently, the resulting sequence, which contains both the transformed patches and position information, is sent to the decoder.

\subsection{Decoder Module}
The decoder performs text recognition using a given image sequence. 
The decoder initially uses the input image sequence and generates the first predicted token by following a beginning token.
This token is a special token named \texttt{[SEP]}, which marks the division between the image and text sequence. 
Thereafter, the model uses the image and predicted token sequence to generate text autoregressively until it reaches the token \texttt{[EOS]}. 
The decoder output is projected by a linear layer from the dimension of the model to the vocabulary size $V$. 
Thereafter, the probabilities are computed on the vocabulary using a softmax function. 
Finally, the beam search is employed to obtain the final output. 
The cross-entropy loss function is used in this process.

The decoder uses GPT~\cite{radford2018gpt, radford2019language} to recognize the text accurately using linguistic knowledge. 
It predicts the next word in a sentence by maximizing the entropy. 
Pre-trained models are publicly available, which eliminates the need for computational resources to acquire language knowledge.

The decoder comprises multiple stacks, with the Transformer layer~\cite{vaswani2017transformer} constituting one block.
This block includes a multi-head mask self-attention and feed-forward network. 
As opposed to previous encoder-decoder structures, this method only uses a decoder for prediction, thereby eliminating the need for cross-attention between the image features and text and significantly simplifying the design.

\subsection{Pre-training with Synthetic Datasets}
The decoder of our method gains language knowledge through GPT in NLP. 
However, it does not connect this knowledge with the image information. 
Thus, we trained the model on artificially generated datasets that included various text forms, such as scenes, handwritten, and printed text, to aid in acquiring image and language knowledge. 
Further details are provided in the experimental section.

\subsection{Fine-Tuning with Real-World Datasets}
Recent studies have demonstrated that synthetic datasets alone are insufficient for handling real-world problems~\cite{baek2021TRBA, bautista2022parseq}. 
Text shapes, fonts, and features may vary depending on the type of recognition required, such as printed or handwritten text.
Therefore, we fine-tuned the pre-trained models using actual data for specific tasks to solve real-world text recognition issues effectively.
The training procedure for the real datasets was the same as that for the synthetic ones.

\subsection{Inference}
The proposed method uses the same training process for inference. 
Patch embedding is employed for the input images and decoder to generate predicted tokens iteratively until the \texttt{[EOS]} token is reached for text recognition.

\section{Experiments} \label{experiments}
\begin{table*}[tb]

\centering
   \caption{
Word accuracy on English scene text recognition benchmark datasets with 36 characters. 
``Synth'' and ``Real'' refer to synthetic and real training datasets, respectively.
}
\label{tab:method_overall_result_english}
\begin{tabular}{cccccccccc}
\toprule
\multirow{3}{*}{Method} &\multirow{3}{*}{Training data} & \multicolumn{8}{c}{Test datasets and \# of samples}                                                                                                                                                                \\ \cline{3-10} 
&                         & IIIT5k                   & SVT                      & \multicolumn{2}{c}{IC13}                            & \multicolumn{2}{c}{IC15}                            & SVTP                     & CUTE                     \\ 
&                         & 3,000                    & 647                      & 857                      & 1,015                    & 1,811                    & 2,077                    & 645                      & 288                      \\ 
\midrule
CRNN~\cite{shi2016crnn} & Synth &81.8 & 80.1 & 89.4 & 88.4 & 65.3 & 60.4 & 65.9 & 61.5 \\ 
$\textrm{ViTSTR}_{\rm BASE}$~\cite{atienza2021vitstr} & Synth & 88.4 & 87.7 & 93.2 & 92.4 & 78.5 & 72.6 & 81.8 & 81.3 \\
TRBA~\cite{baek2021TRBA} & Synth &92.1 & 88.9 & $-$ & 93.1 & $-$ & 74.7 & 79.5 & 78.2 \\
ABINet~\cite{fang2021ABINet} & Synth & 96.2 & 93.5 & 97.4 & $-$ & 86.0 & $-$ & 89.3 & 89.2 \\

PlugNet~\cite{mou2020plugnet} & Synth &94.4 & 92.3 & $-$ & 95.0 & $-$ & 82.2 & 84.3 & 85.0 \\ 
SRN~\cite{yu2020srn} & Synth &94.8 & 91.5 & 95.5 & $-$ & 82.7 & $-$ & 85.1 & 87.8 \\ 
TextScanner~\cite{wan2020textscanner} & Synth &95.7 & 92.7 & $-$ & 94.9 & $-$ & 83.5 & 84.8 & 91.6 \\ 
AutoSTR~\cite{zhang2020autostr} & Synth &94.7 & 90.9 & $-$ & 94.2 & 81.8 & $-$ & 81.7 & $-$ \\ 
PREN2D~\cite{yan2021pren2d} & Synth &95.6 & 94.0 & 96.4 & $-$ & 83.0 & $-$ & 87.6 & 91.7 \\ 
VisionLAN~\cite{wang2021visionlan} & Synth &95.8 & 91.7 & 95.7 & $-$ & 83.7 & $-$ & 86.0 & 88.5 \\ 
JVSR~\cite{bhunia2021joint} & Synth &95.2 & 92.2 & $-$ & 95.5 & $-$ & 84.0 & 85.7 & 89.7 \\ 
CVAE-Feed~\cite{bhunia2021towards} &Synth & 95.2 & $-$ & $-$ & 95.7 & $-$ & 84.6 & 88.9 & 89.7 \\
DiffusionSTR~\cite{fujitake2023diffusionstr} & Synth & 97.3 & 93.6 & 97.1 & 96.4 & 86.0 & 82.2 & 89.2 & 92.5 \\

$\textrm{TrOCR}_{\rm BASE}$~\cite{li2021trocr} & Synth & 90.1 & 91.0 & 97.3 & 96.3 & 81.1 & 75.0 & 90.7 & 86.8  \\
$\textrm{TrOCR}_{\rm LARGE}$~\cite{li2021trocr} & Synth & 91.0 & 93.2 & 98.3 & 97.0 & 84.0 & 78.0 & 91.0 & 89.6  \\

PARSeq~\cite{bautista2022parseq} & Synth& 97.0 & 93.6 & 97.0 & 96.2 & 86.5 & 82.9 & 88.9 & 92.2 \\
$\textrm{MaskOCR}_{\rm BASE}$~\cite{lyu2022maskocr} & Synth & 95.8 & 94.7 & 98.1 & $-$ & 87.3 & $-$ & 89.9 & 89.2 \\
$\textrm{MaskOCR}_{\rm LARGE}$~\cite{lyu2022maskocr} & Synth& 96.5 & 94.1 & 97.8 & $-$ & 88.7 & $-$ & 90.2 & 92.7 \\

$\textrm{SVTR}_{\rm BASE}$~\cite{du2022svtr} & Synth & 96.0 & 91.5 & 97.1 & $-$ & 85.2 & $-$ & 89.9 & 91.7  \\
$\textrm{SVTR}_{\rm LARGE}$~\cite{du2022svtr} & Synth & 96.3 & 91.7 & 97.2 & $-$ & 86.6 & $-$ & 88.4 & 95.1  \\

DTrOCR (Ours) & Synth & \textbf{98.4} & \textbf{96.9} & \textbf{98.8} & \textbf{97.8} & \textbf{92.3} & \textbf{90.4} & \textbf{95.0} & \textbf{97.6} \\
\hline
\hline
CRNN~\cite{shi2016crnn, bautista2022parseq} & Real &94.6 & 90.7 & 94.1 & 94.5 & 82.0 & 78.5 & 80.6 & 89.1 \\ 

TRBA~\cite{baek2021TRBA, bautista2022parseq} & Real &98.6 & 97.0 & 97.6 & 97.6 & 89.8 & 88.7 & 93.7 & 97.7 \\
ABINet~\cite{fang2021ABINet, bautista2022parseq} & Real & 98.6 & 97.8 & 98.0 & 98.0 & 90.2 & 88.5 & 93.9 & 97.7 \\

PARSeq~\cite{bautista2022parseq} & Real& 99.1 & 97.9 & 98.3 & 98.4 & 90.7 & 89.6 & 95.7 & 98.3 \\
DTrOCR (ours) & Real & \textbf{99.6} & \textbf{98.9} & \textbf{99.1} & \textbf{99.4} & \textbf{93.5} & \textbf{93.2} & \textbf{98.6} & \textbf{99.1} \\
%

\bottomrule

\end{tabular}
\end{table*}

\begin{figure}[htp]
	\centering
	\includegraphics[width=1.00\columnwidth, keepaspectratio]{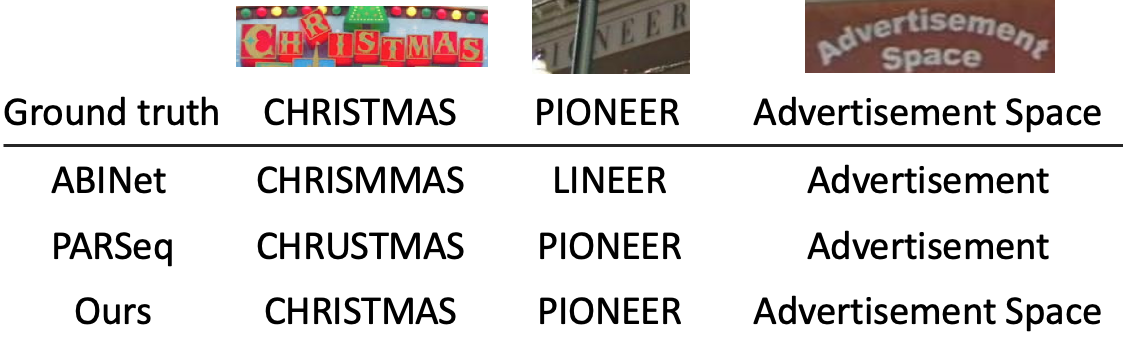}
	\caption{
Comparison of recognition results of state-of-the-art methods and proposed method~\cite{fang2021ABINet, bautista2022parseq}.
The result corresponding to an image is shown on each line, with the ground truth at the top.
The proposed method is robust to occlusion and irregularly arranged scenes and is accurate even for two lines.
}
	\label{fig:recognition_result}
\end{figure}

\begin{table}[ht]
\centering
\caption{Word-level recall, precision and F1 on SROIE Task 2.
}
\label{tab:method_overall_result_sroie}
\begin{tabular}{cccc}
\toprule
Model & Recall & Precision & F1 \\ 
\midrule
CRNN~\cite{shi2016crnn}            & 28.71              & 48.58          & 36.09         \\
H\&H Lab~\cite{huang2019sroie} & 96.35           & 96.52              & 96.43       \\
MSOLab~\cite{huang2019sroie}   & 94.77           & 94.88              & 94.82       \\
CLOVA OCR~\cite{huang2019sroie} & 94.30            & 94.88              & 94.59       \\
$\textrm{TrOCR}_{\rm LARGE}$~\cite{li2021trocr}     & 96.59           & 96.57              & 96.58       \\
\hline
DTrOCR (ours) & \textbf{98.24}            & \textbf{98.51}              & \textbf{98.37}       \\
\bottomrule
\end{tabular}
\end{table}

\begin{table}[ht]
\caption{
    CER on IAM Handwriting Database, where a lower score is better.
}
\label{tab:method_overall_result_iam}
\resizebox{1.0\columnwidth}{!}{
\centering
\begin{tabular}{cccc}
\toprule
Model               & Training Data   & External LM & CER \\
\midrule

Bluche \textit{et al.}~\cite{bluche2017gated}       & Synthetic + IAM          & Yes                  & 3.20          \\
Michael \textit{et al.}~\cite{michael2019evaluating} & IAM                      & No                   & 4.87         \\ 
Wang \textit{et al.}~\cite{wang2020decoupled}    & IAM                      & No                   & 6.40          \\
Kang \textit{et al.}~\cite{kang2022pay}          & Synthetic + IAM          & No                   & 4.67         \\
Diaz \textit{et al.}~\cite{diaz2021rethinking} & Internal + IAM          & Yes                  & 2.75         \\
$\textrm{TrOCR}_{\rm LARGE}$~\cite{li2021trocr}             & Synthetic + IAM          & No                   & 2.89             \\
\hline
DTrOCR (ours)               & Synthetic + IAM          & No                   & \textbf{2.38}             \\
\bottomrule
\end{tabular}
}
\end{table}

\begin{table*}[t]
\caption{
Word accuracy on CTR benchmark.
}
\label{tab:method_overall_result_chinese}
\centering
\scalebox{0.97}{
\begin{tabular}{ccccccr}
\toprule
\multirow{2}*{Method} & \multicolumn{4}{c}{Dataset} & \multirow{2}*{Parameters (M)} & \multirow{2}*{FPS} \\
\cmidrule{2-5}
~  & Scene & Web & Document & Handwriting \\
\midrule
CRNN~\cite{shi2016crnn} & 54.9 & 56.2 & 97.4 & 48.0  & \textbf{12.4} & \textbf{751.0} \\
ASTER~\cite{shi2018aster} & 59.4  & 57.8  & 97.6  & 45.9  & 27.2 & 107.3 \\
MORAN~\cite{luo2019moran} & 54.7 & 49.6 & 91.7 & 30.2 & 28.5 & 301.5 \\
SAR~\cite{li2019show} & 53.8 & 50.5 & 96.2 & 31.0 & 27.8 & 93.1 \\
SEED~\cite{qiao2020seed} & 45.4  & 31.4  & 96.1  & 21.1  & 36.1 & 106.6 \\
MASTER~\cite{lu2021master} & 62.1  &	53.4  & 82.7  & 18.5  & 62.8 & 16.3 \\
ABINet~\cite{fang2021ABINet} & 60.9 & 51.1 & 91.7 & 13.8 & 53.1 & 92.1 \\
TrOCR~\cite{li2021trocr} & 67.8  & 62.7 & 97.9 & 51.7 & 83.9 & 164.6 \\
$\textrm{MaskOCR}_{\rm BASE}$~\cite{lyu2022maskocr} & 73.9  & 74.8 & 99.3 & 63.7 & 100 & $-$ \\
$\textrm{MaskOCR}_{\rm LARGE}$~\cite{lyu2022maskocr} & 76.2  & 76.8 & 99.4 & 67.9 & 318 & $-$ \\
\hline
DTrOCR (ours)  & \textbf{87.4}  & \textbf{89.7} & \textbf{99.9} & \textbf{81.4} & 105 & 97.9 \\
\bottomrule
\end{tabular}
}
\end{table*}

We evaluated the performance of the proposed method on scene image, printed, and handwritten text recognition in English and Chinese.

\subsection{Datasets}
\noindent
\textbf{Pre-training with Synthetic Datasets.}
Our proposed method was pre-trained using synthetic datasets to connect the visual and language information in the LM of the decoder. 
Previous studies~\cite{li2021trocr} obtained training data by extracting available PDFs from the Internet and using real-world receipt data with annotations that are generated by commercial OCR. 
However, substantial time and effort are required to prepare such data.
We created annotated datasets from a text corpus using an artificial generation method to make the process more reproducible. 
We used large datasets that are commonly used to train LMs as our corpus for generating synthetic text images:
PILE~\cite{gao2020pile} for English, CC100~\cite{wenzek2020ccnet}, and the Chinese NLP Corpus\footnote{\url{https://github.com/crownpku/awesome-chinese-nlp}} for Chinese with preprocessing~\cite{zhao2019uer}.

We used three open-source libraries to create synthetic datasets from our corpus. 
We randomly divided the corpus into three categories: scene, printed, and handwritten text recognition, with a distribution of 60\%, 20\%, and 20\%, respectively, to ensure text recognition accuracy. 
We generated four billion horizontal and two billion vertical images of text for scene text recognition using SynthTIGER~\cite{yim2021synthtiger}.
We employed the default font for English and 64 commonly used fonts for Chinese.
We used the Multiline text image configuration in SynthTIGER, set the word count to five, and generated 100 million images using the MJSynth~\cite{jaderberg2016MJSynth} and SynthText~\cite{gupta2016synthtext} corpora for the recognition of multiple lines of English text. 

We created two billion datasets for printed text recognition using the default settings of Text Render\footnote{\url{https://github.com/oh-my-ocr/text_renderer}}. 
Additionally, we employed TRDG\footnote{\url{https://github.com/Belval/TextRecognitionDataGenerator}} to generate another two billion datasets for recognizing handwritten text. 
We followed the methods outlined in previous studies for our process~\cite{li2021trocr}.
We used 5,427 English and four Chinese handwriting fonts~\footnote{\url{https://fonts.google.com/?category=Handwriting}}.

\noindent
\textbf{Fine-Tuning with Real-World and Evaluation Datasets.}
We fine-tuned the pre-trained models on each dataset and evaluated their performance on benchmarks. 
English scene text recognition models have traditionally been trained on large synthetic datasets owing to the limited availability of labeled real datasets.
However, with the increasing amount of real-world datasets, models are now also being trained on real data. 
Therefore, following previous studies~\cite{baek2021TRBA, bautista2022parseq},
we trained our models on both synthetic and real datasets to validate the performance.
Specifically, we used MJSynth~\cite{jaderberg2016MJSynth} and SynthText~\cite{gupta2016synthtext} as synthetic datasets and COCO-Text~\cite{veit2016coco}, RCTW~\cite{shi2017icdar2017}, Uber-Text~\cite{zhang2017uber}, ArT~\cite{chng2019icdar2019}, LSVT~\cite{sun2019icdar}, MLT19~\cite{nayef2019icdarmlt}, and ReCTS~\cite{zhang2019icdar} as real datasets.
Each model was evaluated on six standard scene text datasets: ICDAR 2013 (IC13)~\cite{karatzas2013icdar}, Street View Text (SVT)~\cite{wang2011svt}, IIIT5K-Words (IIIT5K)~\cite{mishra2012iiit}, ICDAR 2015 (IC15)~\cite{karatzas2015icdar}, Street View Text-Perspective (SVTP)~\cite{phan2013svtp}, and CUTE80 (CUTE)~\cite{risnumawan2014cute80}.
The initial three datasets mainly consist of standard text images, whereas the remaining 
datasets include images of text that are either curved or in perspective.  

Thereafter, we tested the accuracy of the printed text recognition in receipt images using Scanned Receipts OCR and Information Extraction (SROIE) Task 2~\cite{huang2019sroie}. 
A total of 626 training and 361 evaluation receipt images were used in the testing.

We employed the widely used IAM Handwriting Database to evaluate the English HWR.
Aachen's partition\footnote{\url{https://github.com/jpuigcerver/Laia}} was used, which resulted in a training set of 6,161 lines from 747 forms, a validation set of 966 lines from 115 forms, and a test set of 2,915 lines from 336 forms.

We evaluated the models for CTR on a large CTR benchmark dataset~\cite{yu2021btcr}.
This dataset includes four subsets (scene, web, document, and handwriting), with a total of 1.4 million fully labeled images.
The scene subset is derived from scene text datasets such as RCTW~\cite{shi2017icdar2017}, ReCTS~\cite{zhang2019icdar}, LSVT~\cite{sun2019icdar}, ArT~\cite{chng2019icdar2019}, and CTW~\cite{yuan2019large}.
It consists of 509,164, 63,645, and 63,646 samples for training, validation, and testing, respectively.
The web subset is built on the MTWI dataset~\cite{he2018MTWI}, with 112,471, 14,059, and 14,059 samples for training, validation, and testing, respectively.
The document subset was generated in document style by Text Render and consists of 400,000 training, 50,000 validation, and 50,000 testing samples.
The handwriting subset was obtained from the handwriting dataset SCUT-HCCDoc~\cite{zhang2020scut}. 
It consists of 74,603, 18,651, and 23,389 training, validation, and testing samples, respectively.

\subsection{Evaluation Metrics}
We used different metrics for the various benchmarks.
The word accuracy was used for the standard scene text recognition and CTR benchmarks; a prediction was considered as correct if the characters at all positions matched. 
SROIE Task 2 was evaluated using the word-level precision, recall, and F1 scores.
Finally, the character error rate (CER) with case sensitivity was used for the HWR benchmark IAM.

We followed the standard protocols for the English scene text recognition~\cite{bautista2022parseq, baek2021TRBA} and CTR~\cite{yu2021btcr} to process the predictions and ground truth.
We filtered the string to fit the 36-character character set (lowercase alphanumeric) to ensure a fair comparison in the English scene text recognition task. 
We implemented specific processes for the CTR: (i) full-width characters were converted into half-width ones; (ii) traditional Chinese characters were converted into simplified ones; (iii) uppercase letters were converted into lowercase ones; and (iv) all spaces were removed.

\subsection{Implementation Details}

We used the English\footnote{\url{https://huggingface.co/gpt2}} and Chinese\footnote{\url{https://huggingface.co/uer/gpt2-chinese-cluecorpussmall}} GPT-2~\cite{radford2019language} models with 12 layers, 768 hidden dimensions, and a Transformer with 12 heads for our decoder model.
These models use a bytepair encoding vocabulary~\cite{sennrich2016bpe}, and we followed previous research~\cite{bautista2022parseq} for image patch embedding with a size of $8 \times 4$. 
We used relative position encoding and set the maximum token length to 512.

We used an English pre-training dataset for English and a combination of English and Chinese datasets for Chinese to train the proposed model. 
Model training was performed for one epoch with a batch size of 32.
We used the AdamW optimizer with a learning rate of 1e-4~\cite{loshchilov2018adamw}.

During the fine-tuning phase, the models were initialized using pre-trained weights and subsequently trained for the target datasets using a learning rate of 5e-6 for one epoch, except for SROIE, for which the models were trained for four epochs.
The same optimizer and batch size were used in the pre-training phase.

We approximately followed previous work for the data augmentation and label pre-processing~\cite{bautista2022parseq}.
We applied RandAugment~\cite{cubuk2020randaugment}, except for Sharpness.
Invert, Gaussian blur, and Poisson noise were added owing to their effectiveness.
The RandAugment policy of three layers and a magnitude of five was used.
All images were resized to $128 \times 32$ pixels.
Furthermore, the original orientation was retained, rotated clockwise, or rotated with a probability of 95\%, 2.5\%, and 2.5\%, to account for images that were rotated 90 degrees clockwise. 
Eventually, the image was standardized to fit into the range of -1 to 1.

The models were trained using PyTorch on Nvidia A100 GPUs with mixed precision. 
The inference was performed on the RTX 2080 Ti to measure the processing speed under the same conditions as those of a previous study~\cite{yu2021btcr}. The reported scores are averaged from four replicates per model, following a previous study~\cite{bautista2022parseq}, except for SROIE Task 2.

\subsection{Comparison with State-of-the-Art Methods}
\noindent
\textbf{Scene Text Recognition.}
We compared the proposed method with several state-of-the-art scene text recognition methods.
Table~\ref{tab:method_overall_result_english} presents the results for several widely used English benchmark datasets: IIIT5K, SVT, IC13, IC15, SVTP, and CUTE.
As the training dataset conditions differed for each method, we present the results of the synthetic and real-world datasets separately.
As can be observed from Table~\ref{tab:method_overall_result_english}, our method outperformed the existing state-of-the-art methods on all benchmarks by a large margin for both the synthetic and real datasets.
The competitive methods for the synthetic datasets, namely TrOCR, ABINet, PARSeq, and MaskOCR, employ an encoder and incorporate LMs to achieve high accuracy.
However, our method, which does not use an encoder, achieved superior accuracy.
This suggests that vision encoders are not necessary to achieve high accuracy in scene text recognition tasks.
As previous studies have demonstrated that training on real datasets is more effective than that on synthetic datasets~\cite{baek2021TRBA}, we also trained our proposed method on real datasets.
We confirmed that the proposed method also achieved better accuracy when it was trained on real datasets.

Figure~\ref{fig:recognition_result} depicts the recognition results of training several state-of-the-art methods~\cite{fang2021ABINet, bautista2022parseq} on real datasets.
The proposed method performed text recognition reasonably well compared to the other methods, even under occlusion and an irregular layout.
Moreover, the proposed method correctly read the two-line text, which methods in previous studies failed to achieve.

\noindent
\textbf{SROIE Task 2.}
Table~\ref{tab:method_overall_result_sroie} presents the results of the existing and proposed methods on SROIE Task 2.
The CRNN, H\&H Lab, MSOLab, and CLOVA OCR methods use CNN-based feature extractors to take advantage of image information, whereas
TrOCR uses the ViT families.
The results demonstrate that our method outperformed the existing methods without either approach.
Thus, the proposed method can be applied not only to reading text in natural scene images but also to text in printed documents in the real world.

\noindent
\textbf{IAM Handwriting Database.}
Table~\ref{tab:method_overall_result_iam} summarizes the results of the proposed and existing methods on the IAM Handwriting Database. 
Our method was superior to the most significant existing approach, which was trained with Diaz's internal annotated dataset and used an external LM~\cite{diaz2021rethinking}. 
Our method does not make use of either of these and can achieve better accuracy solely through synthetic and benchmark datasets.
Our method also performed better than TrOCR, which uses Transformers, under similar conditions. 
The experimental results affirm that the proposed text recognition method with the generative LM is also effective for recognizing handwritten text.

\noindent
\textbf{CTR.}
Table~\ref{tab:method_overall_result_chinese} summarizes the results of the proposed and previous approaches on the CTR benchmark, which has more characters to categorize and is more challenging than the English benchmark. The results confirm the generality of our method.
In terms of accuracy, the proposed method outperformed the existing methods by a large margin for all subsets.
The Transformer-based encoder-decoder methods, namely TrOCR and MaskOCR, achieved high accuracy in previous studies.
Both of those decoders are pre-trained models based on MLM.
However, the proposed method is based on generative pre-training by predicting the next word token.
We confirm that the decoder with the generative model can model sequential patterns more flexibly, even for complex text sequences such as Chinese.

Our method has fewer parameters and is more accurate than the existing large-scale model $\textrm{MaskOCR}_{\rm LARGE}$~\cite{lyu2022maskocr}.
Therefore, our method significantly improves the tradeoff between the number of parameters and accuracy.
Furthermore, the reported processing speed, which is also known as the frames per second (FPS), has been low in previous studies.
Our work is based on generative LMs, and because acceleration research has recently been conducted to make LMs easier to handle~\cite{dettmers2022llm}, we believe that further improvements in terms of speed can be expected by applying these techniques.

\begin{table}[bt]
\centering
\caption{
	Architecture analysis.
}
\label{tab:ablation_architecture_analysis}
\resizebox{1.0\columnwidth}{!}{
\begin{tabular}{c|cc|cc}
\toprule
Model                                & Encoder & Decoder  & STR & CTR  \\ \midrule
Complete model & $-$ & 12 layers (GPT-2~\cite{radford2019language}) & 97.7 & 89.6  \\
Model (a) & 12 layers (ViT~\cite{dosovitskiy2020vit}) & 12 layers (GPT-2~\cite{radford2019language}) & 97.5 & 90.0  \\
TrOCR & 12 layers (ViT~\cite{dosovitskiy2020vit}) & 12 layers (RoBERTa~\cite{liu2019roberta}) & 92.6 & 79.3  \\
\bottomrule
\end{tabular}
}
\end{table}

\begin{table}[!t]
  \caption{
  Effects of training process.
}
\label{tab:ablation_training_process}
  \centering
\resizebox{0.9\columnwidth}{!}{
  \begin{tabular}{lcc}
    \toprule
    &STR &CTR  \\
    \midrule
    Training from scratch             & 61.0 & 43.4  \\
    $+$ pre-trained decoder            & 88.1 & 81.3  \\
    $+$ data augmentation             & 95.3 & 88.9  \\
    $+$ fine-tuning with real datasets & 97.7 & 89.6  \\
  \bottomrule
\end{tabular}
}
\end{table}

\begin{table}[bt]
\centering
\caption{
	Effects of pre-training dataset.
}
\label{tab:ablation_dataset_analysis}
\resizebox{1.0\columnwidth}{!}{
\begin{tabular}{c|cc|cc}
\toprule
Model                                & Dataset amount & Epochs  & STR & CTR   \\ \midrule
Complete model & 100 \% & 1 & 97.7 & 89.6  \\
Model (b)       & 50 \% & 2 & 97.5 & 89.1 \\
Model (c)       & 25 \% & 4 & 96.2 & 85.7 \\
Model (d)       & 25 \% & 1 & 91.4 & 77.9 \\
\bottomrule
\end{tabular}
}
\end{table}

\begin{table}[bt]
\centering
\caption{
 	Effects of pre-trained decoder architecture.
}
\label{tab:ablation_pretrained_decoder}
\resizebox{1.0\columnwidth}{!}{
\begin{tabular}{c|cc|c}
\toprule
Model                                & Decoder & Parameters (M)  & STR   \\ \midrule
Complete model & GPT-2 & 128 & 97.7  \\
Model (e) & GPT-2 Medium & 359 & 97.9  \\
Model (f) & GPT-2 Large & 778 & 98.3  \\
\bottomrule
\end{tabular}
}
\end{table}

\subsection{Detailed Analysis}
The English scene text recognition (STR) and Chinese Text Recognition (CTR) were used to confirm the effectiveness of the proposed method in detail.
We used the real dataset for training and reported the average subset scores.

\noindent
\textbf{Architecture Analysis.}
We analyzed the effects of the model structure on the performance, as indicated in
Table~\ref{tab:ablation_architecture_analysis}.
The configurations show the number of Transformer layers and models used.
Model (a) used a sequence of features that were extracted by the encoder instead of a patch-embedded image sequence.
Therefore, model (a) was expected to produce more sophisticated features because it incorporates an image-specific encoder.
The results show that model (a) achieved slightly higher accuracy in the CTR, whereas the proposed method was superior in the STR.
As many Chinese characters appear similar in CTR, more sophisticated features may be required, resulting in a difference in the accuracy.
However, sufficient accuracy was also achieved with the decoder-only structure, which indicates that an encoder is not always necessary.
We also trained TrOCR, which uses RoBERTa~\cite{liu2019roberta} as the decoder and is pre-trained with MLM.
The comparison of the decoders confirmed that the model that was pre-trained with GPT was superior in the text recognition task.
Thus, the architecture analysis confirms that the text recognition model may not require the encoder-decoder architecture, and GPT is a better decoder approach.

\noindent
\textbf{Effects of Model Training.}
We verified the effects of the training process on the accuracy of our method.
Table~\ref{tab:ablation_training_process} presents the results for each training process.
The decoder initialization using pre-trained models, data augmentation, and fine-tuning with real datasets yielded significant improvements over the training from scratch.

\noindent
\textbf{Effects of Pre-training Dataset.}
We examined the effects of the amount of pre-training datasets and training epochs, as summarized in Table~\ref{tab:ablation_dataset_analysis}.
The proposed method was trained on the entire synthetic dataset for one epoch to avoid overfitting models. 
Models (b) and (c) were trained by reducing the training data using random sampling while the overall number of training iterations was maintained.
The results indicate that increasing the number of data variations is more critical for our method than increasing the training iterations with fewer data.
Furthermore, model (d), which was simply trained with a reduced amount of data, exhibited significantly reduced accuracy, thereby confirming the importance of the pre-training dataset size.

\noindent
\textbf{Model Size Effects of Pre-trained Decoder.}
We verified the variation in the accuracy according to the decoder.
Table~\ref{tab:ablation_pretrained_decoder} presents the number of parameters and accuracy of each decoder.
This experiment was conducted on English benchmarks owing to the limited availability of pre-trained models.
The more extensive models tended to achieve higher accuracy. 
Thus, the experimental results confirm that a higher ability of the LM is also more critical for text recognition.

\section{Conclusion}\label{conclusion}
We have presented a new text recognition method known as DTrOCR, which uses a decoder-only Transformer model.
In this method, a powerful generative LM that is pre-trained on a large corpus is used as a decoder, and the correspondence between the input images and recognition texts is learned.
We demonstrated that in this manner, a simple text recognition structure considering linguistic information is possible.
The experimental results revealed that our method outperformed existing works on various benchmarks.
This study contributes to a fundamental shift in text recognition by showing that text recognition can be performed without the encoder-decoder structure approach and highlighting the possibility of decoder-only Transformer models.

\clearpage

{\small
\bibliographystyle{ieee_fullname}
\bibliography{article}
}

\end{document}